\documentclass[sigconf]{acmart}
\AtBeginDocument{%
  }

\copyrightyear{2025}
\acmYear{2025}
\setcopyright{acmlicensed}\acmConference[KDD '25]{Proceedings of the 31st ACM SIGKDD Conference on Knowledge Discovery and Data Mining V.2}{August 3--7, 2025}{Toronto, ON, Canada}
\acmBooktitle{Proceedings of the 31st ACM SIGKDD Conference on Knowledge Discovery and Data Mining V.2 (KDD '25), August 3--7, 2025, Toronto, ON, Canada}
\acmDOI{10.1145/3711896.3736990}
\acmISBN{979-8-4007-1454-2/2025/08}

\usepackage{multirow}
\usepackage{bbding}
\usepackage{bm}
\usepackage{subfigure}
\usepackage{threeparttable}
\usepackage{enumitem}
\usepackage{xspace}
\usepackage{caption}

\usepackage{balance}

\usepackage{algorithm}
\usepackage{algpseudocode}
\usepackage{placeins}

\makeatletter
\DeclareRobustCommand\onedot{\futurelet\@let@token\@onedot}
\def\@onedot{\ifx\@let@token.\else.\null\fi\xspace}
 
\def\eg{\emph{e.g}\onedot} 
\def\ie{\emph{i.e}\onedot}

\makeatother

\begin{document}

\title{Graph Positional Autoencoders as Self-supervised Learners}


\author{Yang Liu}
\authornote{Both authors contributed equally to this research.}
\orcid{0000-0002-6230-0282}
\affiliation{%
  \institution{Beijing University of Posts and Telecommunications}
  \city{Beijing}
  \country{China}
}
\email{liuyangjanet@bupt.edu.cn}

\author{Deyu Bo}
\authornotemark[1]
\orcid{0000-0003-2063-8223}
\affiliation{%
  \institution{National University of Singapore}
  \country{Singapore}
}
\email{bodeyu1996@gmail.com}

\author{Wenxuan Cao} 
\orcid{0009-0001-0304-8363} 
\affiliation{%
\institution{Beijing University of Posts and Telecommunications} 
\city{Beijing} 
\country{China} 
}
\email{wenxuanc@bupt.edu.cn} 

\author{Yuan Fang}
\orcid{0000-0002-4265-5289}
\affiliation{%
  \institution{Singapore Management University}
  \country{Singapore}
}
\email{yfang@smu.edu.sg}

\author{Yawen Li}
\orcid{0000-0003-2662-3444}
\affiliation{%
  \institution{Beijing University of Posts and Telecommunications}
  \city{Beijing}
  \country{China}
}
\email{warmly0716@126.com}

\author{Chuan Shi}
\authornote{Corresponding author.}
\orcid{0000-0002-3734-0266}
\affiliation{%
  \institution{Beijing University of Posts and Telecommunications}
  \city{Beijing}
  \country{China}
}
\email{shichuan@bupt.edu.cn}
\renewcommand{\shortauthors}{Yang Liu et al.}

\begin{abstract}
Graph self-supervised learning seeks to learn effective graph representations without relying on labeled data. 
Among various approaches, graph autoencoders (GAEs) have gained significant attention for their efficiency and scalability.
Typically, GAEs take incomplete graphs as input and predict missing elements, such as masked node features or edges.
Although effective, our experimental investigation reveals that traditional feature or edge masking paradigms primarily capture low-frequency signals in the graph and fail to learn expressive structural information.
To address these issues, we propose Graph Positional Autoencoders (GraphPAE), which employ a dual-path architecture to reconstruct both node features and positions.
Specifically, the feature path uses positional encoding to enhance the message-passing processing, improving the GAEs' ability to predict the corrupted information. 
The position path, on the other hand, leverages node representations to refine positions and approximate eigenvectors, thereby enabling the encoder to learn diverse frequency information.
We conduct extensive experiments to verify the effectiveness of GraphPAE, including heterophilic node classification, graph property prediction, and transfer learning. 
The results demonstrate that GraphPAE achieves state-of-the-art performance and consistently outperforms the baselines by a large margin.

\end{abstract}

\ccsdesc[500]{Information systems~Data mining}

\keywords{Graph Neural Networks, Self-supervised Learning, Graph Autoencoders, Positional Encoding}


\maketitle

\newcommand\kddavailabilityurl{https://doi.org/10.5281/zenodo.15516721}

\ifdefempty{\kddavailabilityurl}{}{
\begingroup\small\noindent\raggedright\textbf{KDD Availability Link:}\\
The source code of this paper has been made publicly available at \url{\kddavailabilityurl}.
\endgroup
}

\section{Introduction}

\begin{table}
\caption{Comparison of different graph autoencoders.}
  \resizebox{\linewidth}{!}{
  \begin{tabular}{lcccccc}
\toprule
\multirow{3}{*}{Model} & \multicolumn{3}{c}{Corruption} & \multicolumn{3}{c}{Reconstruction} \\ 
\cmidrule(lr){2-4} 
\cmidrule(lr){5-7} 
                       & Feature & Edge & Position                       & Feature & Edge & Other \\
\midrule
GraphMAE~\cite{hou2022graphmae} & \Checkmark & - & - & \Checkmark & - & - \\
StructMAE~\cite{10.24963/ijcai.2024/241} & \Checkmark & - & - & \Checkmark & - & - \\
AUG-MAE~\cite{wang2024rethinking} & \Checkmark & - & - & \Checkmark & - & - \\
S2GAE~\cite{tan2023s2gae} & - & \Checkmark & - & - & \Checkmark & - \\
SeeGera~\cite{li2023seegera} & \Checkmark & \Checkmark & - & \Checkmark & \Checkmark & - \\
Bandana~\cite{zhao2024masked} & - & \Checkmark & - & - & \Checkmark & - \\
MaskGAE~\cite{li2023s} & - & \Checkmark & - & \Checkmark & - & Degree \\
GiGaMAE~\cite{shi2023gigamae} & \Checkmark & \Checkmark & - & - & - & Latent \\
\midrule
Our work & \Checkmark & - & \Checkmark & \Checkmark & - & Position \\
\bottomrule
\end{tabular}}
\label{tab:comparision}
\end{table}

\begin{figure*}[t]
    \centering
    \label{fig:frequency}
    \subfigure[Feature Masking]{
        \centering
        \includegraphics[width=0.32\linewidth]{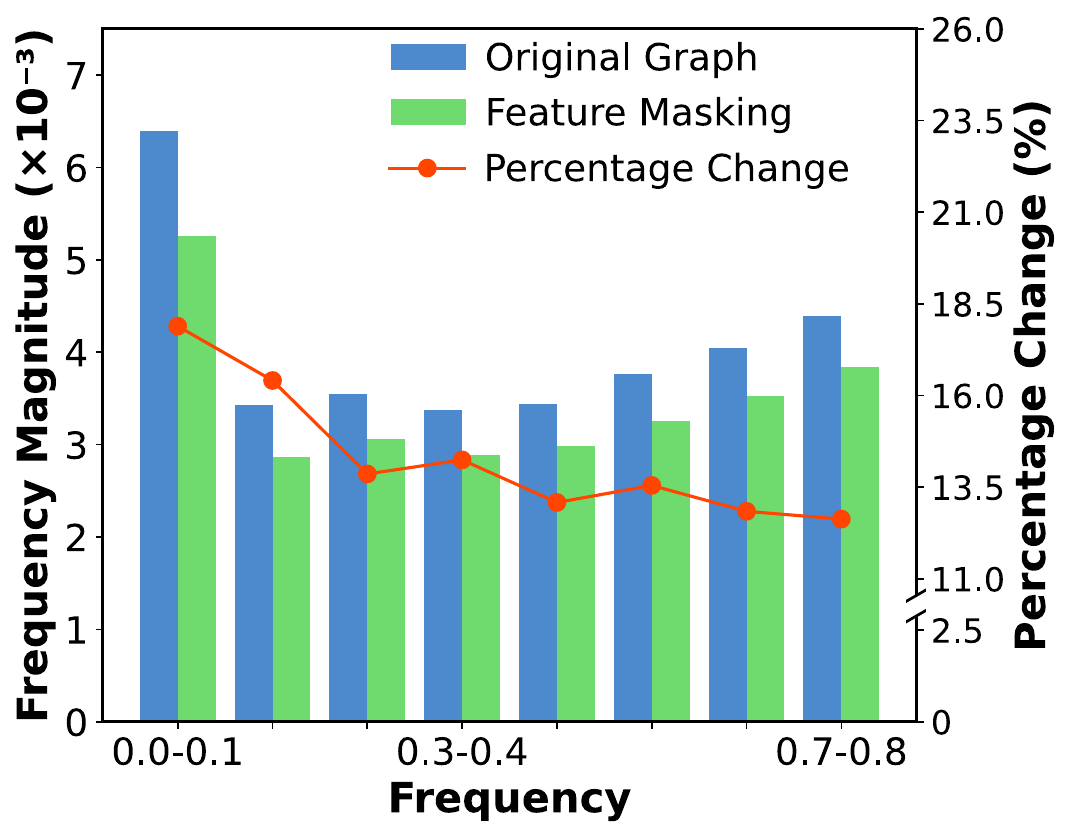}
        \label{fig:mask_feat}
    }
    \subfigure[Edge Masking]{
        \centering
        \includegraphics[width=0.32\linewidth]{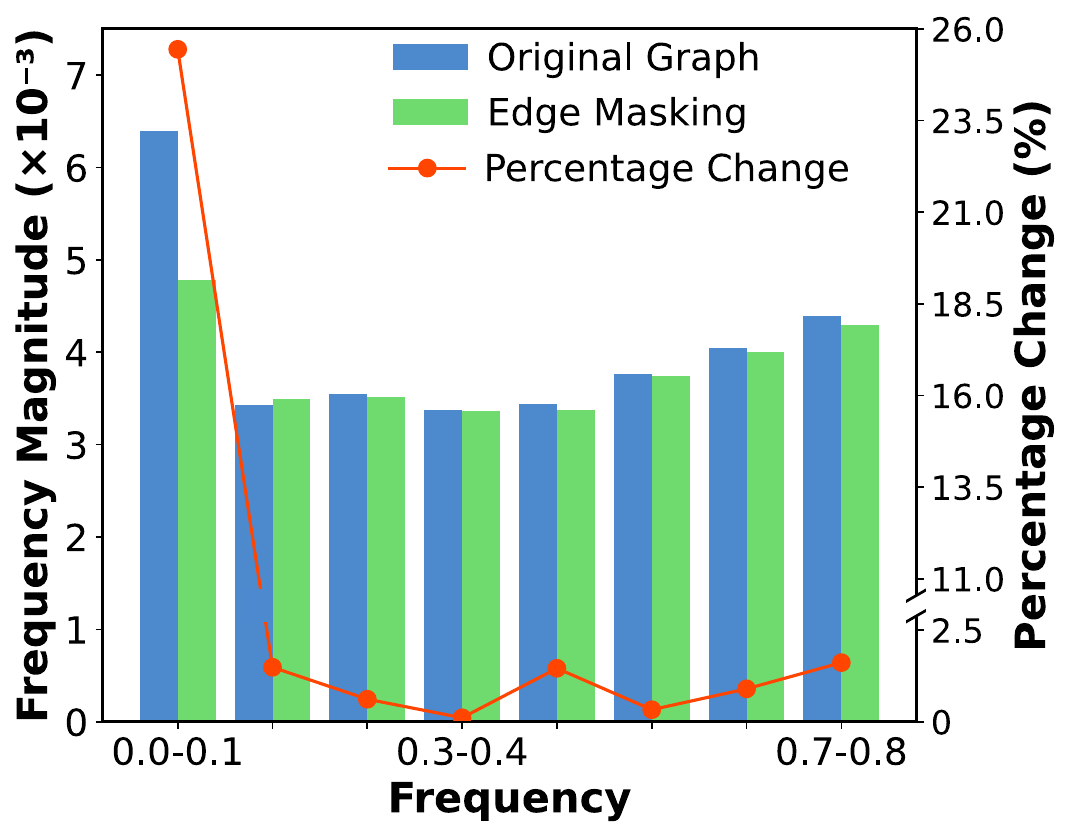}
        \label{fig:mask_edge}
    }
    \subfigure[Eigenvector offsetting]{
        \centering
        \includegraphics[width=0.32\linewidth]{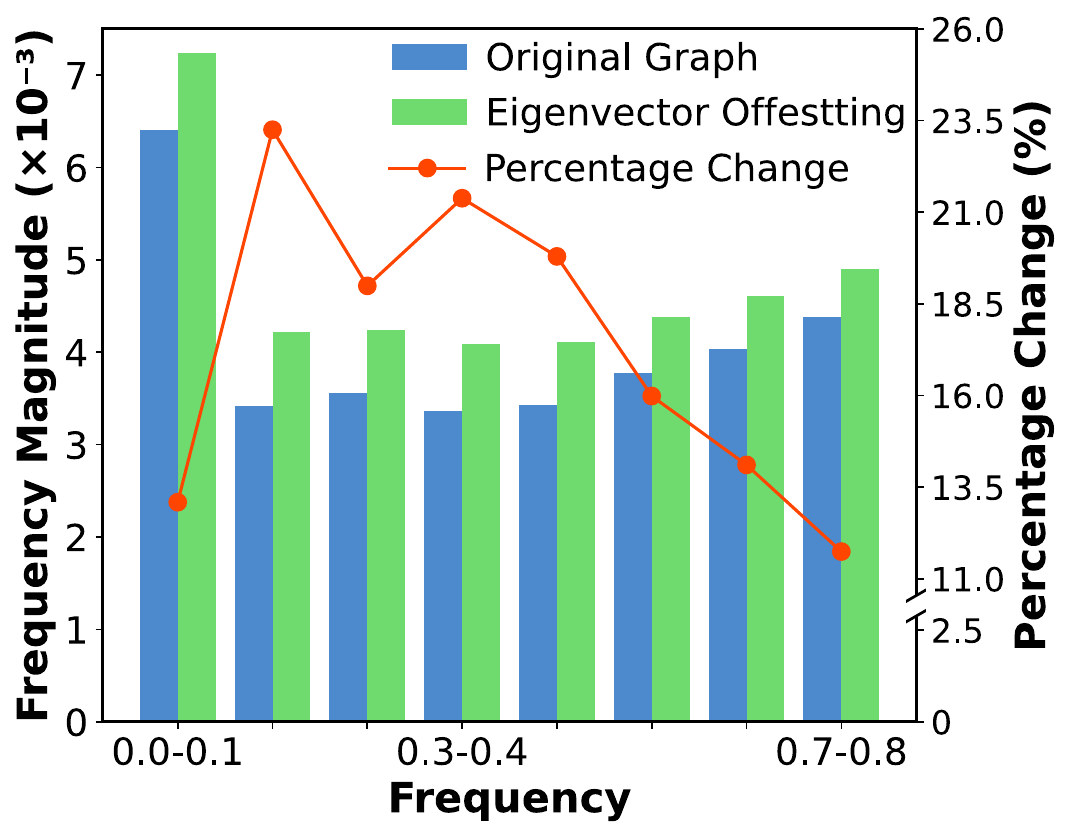}
        \label{fig:position offset}
    }
    \caption{Frequency magnitudes of the original and corrupted graphs in the Squirrel dataset.}
\end{figure*}

Graph neural networks (GNNs) have achieved significant success across various fields, including social network analysis~\cite{carrington2005models, singh2024social, yu2025gcot}, recommendation systems~\cite{gao2023survey, sharma2024survey}, and drug discovery~\cite{jiang2021could, bongini2021molecular, xiong2019pushing}.
However, training effective GNNs in real-world applications remains challenging due to the limited availability of labeled data in many domains~\cite{yu2024few, zhu2025graphclip}.
To address this problem, graph self-supervised learning is proposed to learn graph representations without labeled data~\cite{liu2022graph, xie2022self, yu2024generalized}.

Among existing approaches, the generative~\cite{wu2021self, waikhom2023survey} and contrastive~\cite{ju2024towards} learning paradigms have dominated recent advances. In particular, graph autoencoders (GAEs)~\cite{kipf2016variational, hou2022graphmae} have gained attention due to their simplicity, efficiency, and scalability. The GAEs follow a corruption-reconstruction framework, which learns graph representations by recovering the missing information of the incomplete input graphs, as Table \ref{tab:comparision} illustrates.
For example, GraphMAE~\cite{hou2022graphmae} replaces node features with a learnable token, while Bandana~\cite{zhao2024masked} proposes a non-discrete edge masking strategy.
Moreover, some GAEs even go beyond reconstructing node features and edges by targeting structural features like degree~\cite{li2023s}.
Despite their success, the performance of GAEs highly depends on the choice of corruption and reconstruction objectives. Therefore, a natural question arises:
\textit{Do existing feature or edge masking mechanisms fully exploit the graph data? If not, what alternative objectives could be designed to further improve the performance of GAEs?}
A well-informed answer can help us identify the weaknesses of existing masking strategies and deepen our understanding of GAEs.

To answer this question, we first revisit the feature and edge masking strategies from a spectral perspective~\cite{bo2023survey, bo2023specformer}.
Specifically, we transform the node features $\mathbf{X}$ into the spectral domain by using the eigenvectors of graph Laplacian $\mathbf{U}^{\top}$ and examine its frequency magnitude $\mathbf{U}^{\top}\mathbf{X}$.
In case of feature masking, we randomly mask 20\% node features and denote it as $\widetilde{\textbf{X}}$, while for edge masking, we randomly remove 20\% edges and construct the corrupted graph eigenvectors $\widetilde{\textbf{U}}^{\top}$.
The frequency magnitude changes of $\mathbf{U}^{\top}\widetilde{\textbf{X}}$ and $\widetilde{\textbf{U}}^{\top}\mathbf{X}$ are shown in Figures~\ref{fig:mask_feat} and~\ref{fig:mask_edge}, respectively.
It can be observe that the magnitude differences between original and corrupted graphs are more pronounced in the low-frequency band, \ie, [0.0, 0.1], while the differences decrease at higher frequencies.
Notably, to effectively minimize the reconstruction loss, GAEs will primarily focus on the frequency bands with larger discrepancies.
As a result, existing GAEs mainly reconstruct low-frequency information and overlook the high-frequency information, which has been shown to be valuable in real-world tasks~\cite{bo2021beyond, wang2020gcn}.
More experimental details are provided in Appendix \ref{app:freq}.

Given the above weakness of existing GAEs, it is natural to ask: \textit{How can we design the corruption and reconstruction objectives to exploit the diverse frequency information?}
Essentially, the eigenvectors of graph Laplacian represent different frequencies.
Directly perturbing and reconstructing the eigenvectors can help GAEs learn different frequency information in graphs.
Figure \ref{fig:position offset} illustrates the frequency magnitudes of the eigenvectors with random offsets, \ie, $\widetilde{\mathbf{U}}^{\top}X=(\mathbf{U}+\delta)^{\top}X$.
We observe that this strategy perturbs a broader range of frequency components, particularly in higher-frequency, \ie [0.2, 0.5], enabling the model to capture more diverse frequency information.
However, reconstructing the eigenvectors is not a trivial task and presents two main challenges:
(1) Eigenvectors represent the global structural patterns of a graph, which cannot be easily approximated by basic GNNs.
Existing GAEs typically adopt message-passing neural networks (MPNNs) as the encoder, whose expressiveness is bounded by the 1-WL test~\cite{xu2018powerful}. The intrinsic weakness of MPNNs restricts GAEs' ability to capture long-range dependencies between nodes and higher-order graph patterns~\cite{wu2021representing, dwivedi2022long}.
(2) Eigenvectors suffer from sign- and basis-ambiguity issues~\cite{lim2022sign}. Directly reconstructing the eigenvectors leads to non-unique solutions, affecting the robustness of GAEs~\cite{wang2022equivariant, bo2024graph}.

To overcome these challenges, we propose Graph Positional Autoencoders (GraphPAE) for graph self-supervised learning.
GraphPAE adopts a dual-path architecture to address both the expressivity and ambiguity issues.
Specifically, in the feature path, GraphPAE integrates positional encoding (PE) into the message-passing process to enhance the expressiveness of MPNNs, thus improving the model's ability to reconstruct corrupted information.
Moreover, in the position path, node representations are used to refine the PE, which approximates the eigenvectors, thereby transferring diverse frequency information into the encoder.
Finally, in the reconstruction stage, instead of directly recovering the raw eigenvectors, GraphPAE uses the relative node distance as a surrogate objective, avoiding potential ambiguity issues.


\vspace{5pt}
\noindent \textbf{Contributions.} The contributions of our paper are as follows:
\begin{enumerate}[leftmargin=*]
    \item We are the first to explore the masking strategy in GAEs from a spectral perspective. By comparing the frequency magnitudes between original and corrupted graphs, we identify that existing GAEs focus on reconstructing the low-frequency information of graphs and neglecting other frequencies.
    \item We propose GraphPAE, a novel graph positional autoencoder that learns graph representations by reconstructing both node features and positions, thereby enabling GAEs to capture a broader range of frequency information.
    \item We benchmark GraphPAE against state-of-the-art baselines across various tasks, including node classification, graph property prediction, and transfer learning. The results on 14 graphs demonstrate that GraphPAE consistently outperforms the baselines and shows impressive performance on heterophilic graphs.
\end{enumerate}

\section{Related Work}
We provide an overview of two main areas related to our work, namely generative approaches in graph self-supervised learning and positional encoding techniques to enhance structural expressiveness in GNNs.

\subsection{Generative Graph Self-supervised Learning}
Generative graph learning encompasses a set of graph self-supervised learning techniques aimed at reconstructing missing information in incomplete input graphs. It can be broadly categorized into autoregressive and autoencoding approaches.

\vspace{2pt}
\noindent \textbf{Graph Autoregressive models (GARs).} GARs treat sequential graph generation as the pre-training task, where the node or edge is predicted based on its prior context. GPT-GNN~\cite{hu2020gpt} factorizes each node generation into attribute and edge generation and replenishes the omitted parts by an adaptive queue. MGSSL~\cite{zhang2021motif} introduces motif generation into pre-training. GraphsGPT~\cite{gaograph} transforms non-Euclidean graphs into learnable Euclidean words and reconstructs the original graph from these words. 

\vspace{2pt}
\noindent \textbf{Graph Autoencoders (GAEs).} GAEs reconstruct all desired content once from the latent representation output by the encoder. Early GAEs (\emph{e.g.}, VGAE~\cite{kipf2016variational} and ARGVA~\cite{pan2018adversarially}) learn representations through link reconstruction and spark a series of work, including feature reconstruction (\emph{e.g.}, GALA~\cite{park2019symmetric} and WGDN~\cite{cheng2023wiener}) and combination reconstruction of structure and feature (\emph{e.g.}, GATE~\cite{salehi2019graph}). However, these traditional GAEs often perform poorly in downstream tasks, except link prediction, which is attributed to their overemphasis on proximity information at the expense of structural information~\cite{hassani2020contrastive}. 
Recently, masked GAEs~\cite{hustrategies, hou2022graphmae} have become highly successful models for representation learning. GraphMAE~\cite{hou2022graphmae} and GraphMAE2~\cite{hou2023graphmae2} successfully bridge the performance gap between graph contrastive learning and generative learning by reconstructing masked features for training. 
AUG-MAE~\cite{wang2024rethinking} introduces an adversarial masking strategy to enhance feature alignment and add a uniformity regularizer to promote high-quality graph representations.
Additionally, some works are focusing on masking and reconstructing graph structures (\emph{e.g.}, edges~\cite{li2023s, tan2023s2gae} and paths~\cite{li2023s}). Bandana~\cite{zhao2024masked} uses continuous and dispersive edge masks and bandwidth prediction instead of discrete edge masks and reconstruction. Besides, some works mask features and edges simultaneously~\cite{li2023seegera, shi2023gigamae} and propose novel reconstruction objectives such as latent embeddings~\cite{shi2023gigamae}.

\subsection{Graph Positional Encoding}
Positional encodings (PEs), originally designed to enhance transformers by encoding positional information in sequential data, have been introduced in graph learning to provide explicit position-aware features. By assigning unique identifiers to nodes based on their structural positions, PEs enable graph neural networks (GNNs) to distinguish non-isomorphic structures beyond the limitations of the 1-WL test. Existing graph PEs can be divided into two main categories: Laplacian-based PEs and Distance-based PEs.

\vspace{2pt}
\noindent \textbf{Laplacian-based PEs.} These methods leverage the eigenvectors of the graph Laplacian matrix as initial node positional features~\cite{dwivedi2020generalization, dwivedi2023benchmarking, dwivedi2021graph}. Eigenvectors form a natural basis for encoding graph structure, similar to how sinusoidal functions represent positional information in sequential models~\cite{vaswani2017attention}. Serving as PEs, eigenvectors have two traditional constraints, \emph{i.e.}, sign- and basis ambiguity~\cite{lim2022sign}. 
To keep the models robust to sign ambiguity,~\cite{dwivedi2020generalization, dwivedi2023benchmarking} randomly flip the sign of the eigenvectors during training. SAN~\cite{kreuzer2021rethinking} introduces the information of eigenvalues into PEs and leverages a transformer-based positional encoder, enabling models to be more informative and expressive. PEG~\cite{wang2022equivariant} solves the sign- and basis-ambiguity by treating the distance of eigenvectors between node pairs as PEs.~\cite{lim2022sign} propose SignNet and BasisNet to learn sign- and basis-invariant PEs, respectively.~\cite{huang2023stability, bo2024graph} study stable PEs that are robust to disturbance of the Laplacian matrix.

\vspace{2pt}
\noindent \textbf{Distance-based PEs.} These approaches assign positional features to nodes based on their relative distances within the graph structure, which are typically derived from spatial relationships~\cite{rampavsek2022recipe, li2020distance}. One common approach is to use the random walk matrix capturing structural relationships in a probabilistic manner~\cite{dwivedi2021graph, dwivedi2023benchmarking}. Another popular method is to encode the shortest distance between node pairs~\cite{ying2021transformers, rampavsek2022recipe, li2020distance}. For example, Graphormer~\cite{ying2021transformers} incorporates shortest-path distances into its attention mechanism. Additionally, GraphiT~\cite{mialon2021graphit} encodes PEs with a diffusion kernel, enabling more flexible and adaptive representations of node relationships.~\cite{shiv2019novel} introduces a novel positional encoding scheme that extends transformers to tree-structured data, enabling efficient and parallelizable sequence-to-tree, tree-to-sequence, and tree-to-tree mappings.~\cite{wijesinghe2025graph} first introduces PEs in graph self-supervised learning and designs a GNN framework that integrates a k-hop message-passing mechanism to enhance its expressiveness.

\begin{figure*}[t]
    \subfigure[Overall pipeline of GraphPAE.]{
        \centering
    \includegraphics[height=0.52\linewidth]{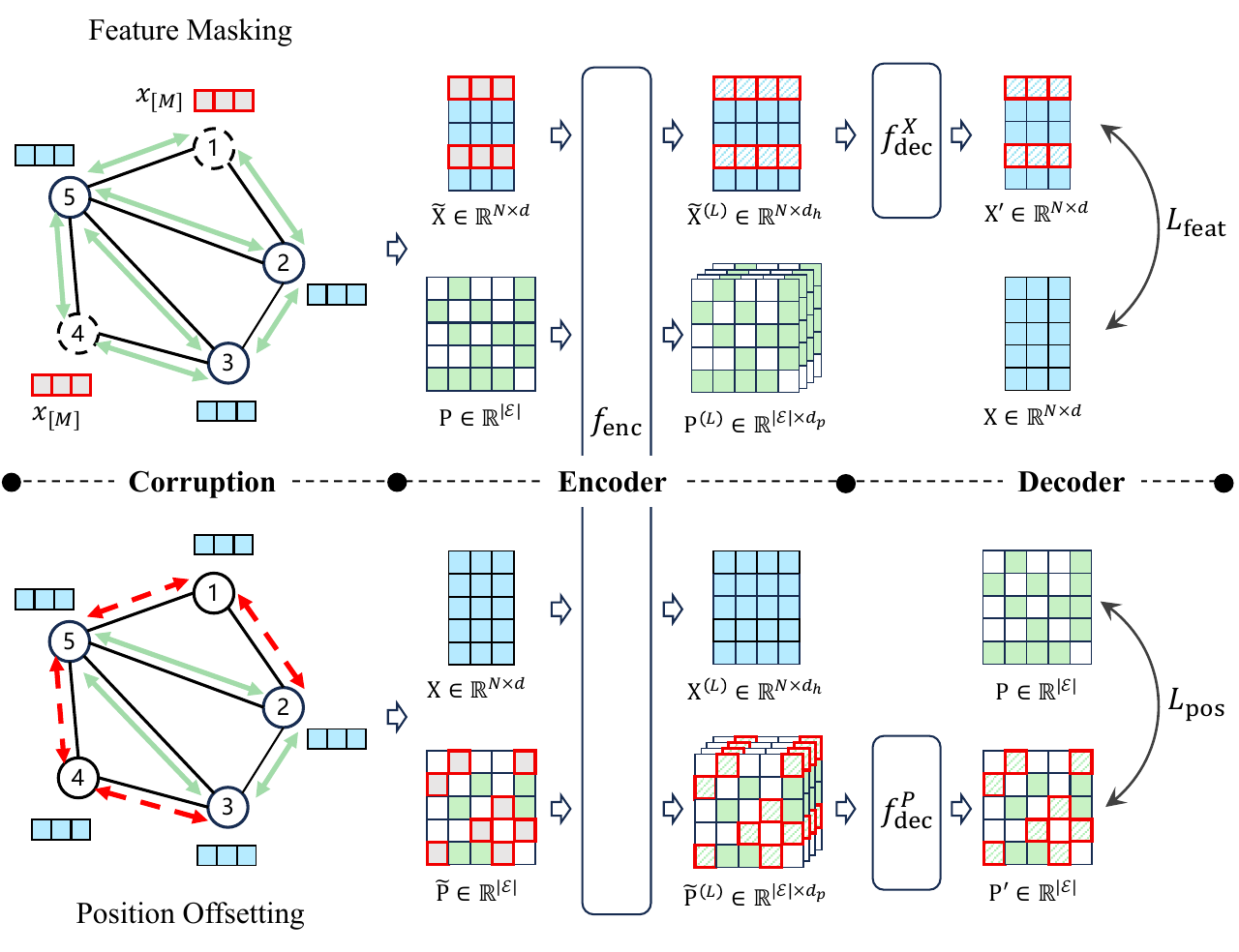}
        \label{fig:overview}
    }\hspace{2mm}
    \subfigure[Encoder of GraphPAE]{
        \centering
    \includegraphics[height=0.52\linewidth]{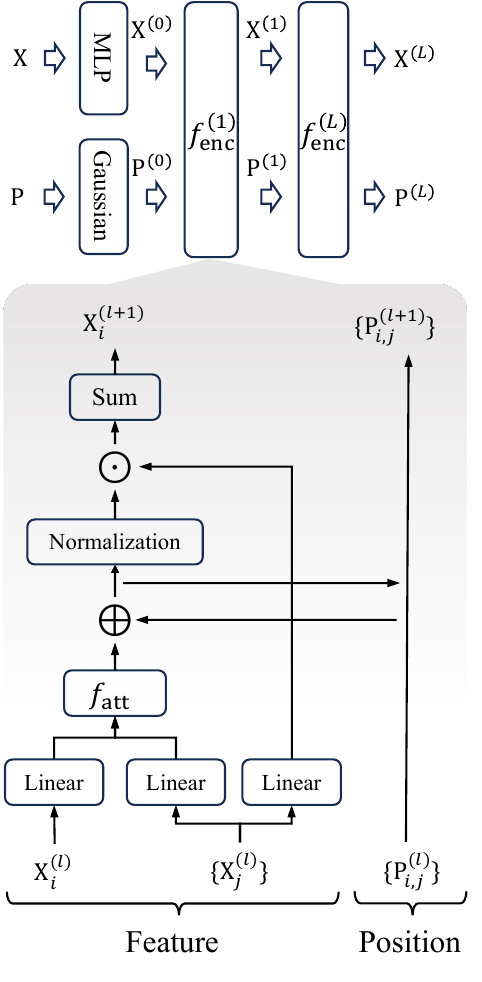}
        \label{fig:encoder}
    }
    \caption{(a): GraphPAE integrates a positional corruption-reconstruction mechanism to encourage the GAE to capture diverse frequency information. 
    For feature reconstruction, masked features are encoded with $\mathbf{P}$ and decoded to recover original features.
    For positional reconstruction, noise is added to $\widetilde{\mathbf{U}}$ to corrupt relative distances, which are then encoded with $\mathbf{X}$ and decoded to recover the original distances.
    Figure (b): The encoder employs a dual-path architecture to update node $\mathbf{X}^{\left(l\right)}$ and positional representations $\mathbf{P}^{\left(l\right)}$ at each layer.
    The feature path integrates positional encodings to enhance message passing, improving the GAE’s ability to reconstruct corrupted features. The position path utilizes node representations to refine positional embeddings to approximate original pairwise distances.}
\end{figure*}

\section{Preliminaries}
Before describing our method, we first define some notations and introduce some important concepts used in this paper.

\vspace{2pt}
\noindent \textbf{Problem Definition.} Given a graph $\mathcal{G}=\{\mathcal{V}, \mathcal{E}, \mathbf{X}\}$, where $\mathcal{V}$ is the node set with $\mathcal{V}=\{v_i\}_{i=1}^{N}$, $\mathcal{E}$ is the set of edges, and $\mathbf{X} \in \mathbb{R}^{N \times d}$ represents the $d$-dimensional node feature matrix. We use the adjacency matrix $\mathbf{A} \in \{0, 1\}^{N \times N}$ to describe the graph structure, where $A_{ij}=1$ if there is an edge between $v_i$ and $v_j$, and $A_{ij} =0$ otherwise. The goal of our framework is to learn an effective graph encoder $f_{enc}(\cdot)$ without relying on labels from downstream tasks. Once trained, the encoder $f_{enc}(\cdot)$ generates node representations as $\mathbf{H} = f_{enc}(\mathcal{G}) \in \mathbb{R}^{N \times d_h}$ or graph representations as $\mathbf{H}_\mathcal{G} = \text{READOUT}\left (\mathbf{H}_{v_i} \mid v_i\in\mathcal{V}\right ) \in \mathbb{R}^{d_h}$, where READOUT is a permutation-invariant function such as mean, max or sum pooling. The representations can then be used to train a simple linear classifier on labeled data from downstream tasks.

\vspace{2pt}
\noindent \textbf{Laplacian Graph Eigenvectors.} The normalized graph Laplacian matrix $\mathbf{L}$ is defined as $\mathbf{L} = \mathbf{I}_{N} - \mathbf{D}^{-1/2} \mathbf{A} \mathbf{D}^{-1/2}$, where $\mathbf{I}_{N}$ is an identity matrix and $\mathbf{D}$ is the degree matrix with $D_{ii}=\sum_{j} A_{ij}$ for $v_i \in \mathcal{V}$, and $D_{ij} = 0$ for $i \neq j$. The Laplacian matrix $\mathbf{L}$ can be decomposed as $\mathbf{L} = \mathbf{U} \bm{\Lambda} \mathbf{U}^{\top}$, where $\bm{\Lambda}=\text{diag}(\{\lambda_{i}\}_{i=1}^{N})$ is the diagonal matrix of eigenvalues, and $\mathbf{U}=[\mathbf{u}_{1}, \cdots, \mathbf{u}_{N}] \in \mathbb{R}^{N \times N}$ consists of a set of eigenvectors. Each eigenvector $\mathbf{u}_{i} \in \mathbb{R}^{N}$ corresponds to eigenvalue $\lambda_{i}$, where the eigenvalues are ordered as $0 \leq \lambda_{1} \leq \cdots \leq \lambda_{N} \leq 2$. Given eigen-decomposition $\mathbf{L}\mathbf{u}_{i} = \lambda_{i}\mathbf{u}_{i}$ with $\mathbf{u}_{i}^\top\mathbf{u}_{i}=1$, we have $\mathbf{u}_{i}^\top\mathbf{L}\mathbf{u}_{i} = \lambda_{i}$. Since $\mathbf{u}_{i}^\top\mathbf{L}\mathbf{u}_{i} = \sum_{(j,k)\in \mathcal{E}} (\mathbf{u}_{i,j} - \mathbf{u}_{i,k})^2$, we finally derive that $\sum_{(j,k)\in \mathcal{E}} (\mathbf{u}_{i,j} - \mathbf{u}_{i,k})^2 = \lambda_{i}$, which indicates that $\lambda_{i}$ reflects the frequency magnitude of eigenvector $\mathbf{u}_{i}$ over the graph. 
In this paper, we follow the popular positional encoding methods and adopt top-$k$ eigenvectors as initial positions to reduce complexity. Therefore, we redefine $\mathbf{U}=[\mathbf{u}_{1}, \cdots, \mathbf{u}_{K}] \in \mathbb{R}^{N \times K}$, where $K \le N$. 

\section{Proposed Framework: GraphPAE}
In this section, we introduce GraphPAE, a novel GAE designed to reconstruct both node features and positions.
There are three key components in GraphPAE: corruption, encoder, and decoder, as illustrated in Figure \ref{fig:overview}.

\subsection{Data Corruption}
\label{session:corruption}

The corruption-reconstruction paradigm has become a basic component of autoencoders.
A well-designed masking strategy can prevent information leakage and enhance model efficiency~\cite{MAE}.
Existing GAEs commonly apply the random masking strategy to the discrete nodes and edges. However, node positions, such as eigenvectors, are continuous by nature and are not suitable for the random masking approach.
Therefore, we adopt different corruption strategies for these two modalities.

\vspace{2pt}
\noindent \textbf{Feature Masking.}
We follow GraphMAE~\cite{hou2022graphmae} and replace the masked node features with a learnable vector. Specifically, we randomly sample a subset of nodes $\widetilde{\mathcal{V}} \subset \mathcal{V}$ and reset their features. The corrupted feature matrix $\widetilde{\mathbf{X}}$ is defined as:
\begin{equation}
    \widetilde{\mathbf{X}}_i = 
    \begin{cases}
    \textbf{x}_{[M]}, & \text{if }v_i \in \widetilde{\mathcal{V}} \\
    \mathbf{X}_i, & \text{if }v_i \notin \widetilde{\mathcal{V}}
    \end{cases}
\end{equation}
where $\mathbf{X}_i \in \mathbb{R}^d$ is the original feature of node $v_i$, and $\textbf{x}_{[M]} \in \mathbb{R}^d$ is a learnable vectors.

\vspace{2pt}
\noindent \textbf{Position Offsetting.}
Inspired by the recent advances in molecular representation learning~\cite{UniMol}, we propose to add random offsets to node positions.
Specifically, we use the top-$k$ eigenvectors of graph Laplacian $\textbf{U} \in \mathbb{R}^{N \times K}$ as the initialization of node positions. Similar to the feature corruption, we define the corruption of node positions as:
\begin{equation}
    \widetilde{\mathbf{U}}_i = 
    \begin{cases}
    \mathbf{U}_i + \delta, & \text{if }v_i \in \widetilde{\mathcal{V}} \\
    \mathbf{U}_i, & \text{if }v_i \notin \widetilde{\mathcal{V}} \\
    \end{cases}
    \label{eq:position_offset}
\end{equation}
where $\mathbf{U}_i \in \mathbb{R}^{K}$ denotes the position for node $v_i$ and $\delta \in \mathbb{R}^K$ is a noise vector sampled from a uniform distribution $\mathcal{U}\left (-\mu_p, \mu_p \right )$. In practice, we set $\mu_p$ to 0.001 or 0.01 for different datasets.

\vspace{2pt}
\noindent \textbf{Relative Positional Encoding.}
Reconstructing eigenvectors enables GAEs to learn different frequency information. However, it is challenging to recover the corrupted eigenvectors as they suffer from the sign- and basis-ambiguity~\cite{lim2022sign,bo2024graph}.
To address this, we compute the Euclidean distance between each pair of nodes to obtain their relative node distances
\begin{equation}
    {\mathbf{P}_{i,j}} = 
\begin{cases}
\|\mathbf{U}_i - \mathbf{U}_j\|_2, & \text{if }\textbf{A}_{i,j}=1 \\
0, & \text{otherwise}
\end{cases}
\label{eq:euclidean_distance}
\end{equation}
where $\mathbf{P} \in \mathbb{R}^{N \times N}$ is the relative distance matrix and we use $\widetilde{\mathbf{P}}$ to indicate its corrupted version.
The relative distance matrix is used as a surrogate of node positions to eliminate the ambiguity of eigenvectors.

\subsection{GraphPAE Encoder}
\label{session:encoder}

After corruption, the node features and pair-wise distances are then fed into the encoder to learn node and position representations through message-passing, which can be formulated as
\begin{equation}
    \mathbf{X}_i^{\left( l+1 \right)}, \mathbf{P}_{i}^{\left( l+1 \right)} = f_{\text{enc}}^{\left (l+1 \right )} \left ( \mathbf{X}_i^{\left (l \right )}, \left \{\mathbf{X}_j^{\left (l \right )}\right \}_{j \in \mathcal{N}_i}, \mathbf{P}_{i}^{\left(l \right)} \right),
\label{eq:node_representations}
\end{equation}
where $l\in\{0, 1, 2, ..., L\}$ indicates the layer of encoder, and $\mathcal{N}_i$ is the neighbors of node $v_i$.
The first layer of the encoder, \eg, $f_{\text{enc}}^{(0)}$, is designed to align the dimensions of node features and position representations.
Specifically, it first lifts the scalar relative distance $\mathbf{P}_{i,j} \in \mathbb{R}$ into a vector representation $\mathbf{P}_{i,j}^{(0)} \in \mathbb{R}^{d_h}$ through a series of Gaussian RBF kernels
\begin{equation}
    \mathbf{P}_{i,j}^{(0)} = \text{MLP} \left( \left[ G(\mathbf{P}_{i,j}; \mu_1, \sigma), \cdots, G(\mathbf{P}_{i,j}; \mu_d, \sigma) \right] \right), 
\label{eq:distance_feature}
\end{equation}
where MLP stands for Multi-layer Perceptron and $G(\mathbf{P}_{i, j}; \mu_k, \sigma) = \exp \left( -\left (\mathbf{P}_{i, j} - \mu_k\right)^2/2\sigma^2 \right)$ is the $k$-th Gaussian basis functions with mean $\mu_k$ and standard deviation $\sigma$.
As for the node features, it uses another MLP to transform them into $d_h$-dimension representations
\begin{equation}
    \mathbf{X}_i^{\left (0 \right )} = \text{MLP}\left( \mathbf{X}_i \right).
\end{equation}
After transformation, the encoder needs to aggregate information from both sides to update the node and position representations layer by layer, as shown in Figure \ref{fig:encoder}.

\vspace{2pt}
\noindent \textbf{Feature Path: PE-enhanced MPNNs.}
It is well-established that the expressive power of traditional MPNNs is bounded by the 1-WL test~\cite{dwivedi2023benchmarking, canturkgraph}.
Fortunately, existing methods prove that adding position information to the message-passing process can significantly improve the expressive power of MPNNs~\cite{dwivedi2021graph, huang2023stability}.
Inspired by the recent progress in graph PEs~\cite{wangequivariant}, we propose to incorporate the position representation into MPNNs as follows:
\begin{equation}
\begin{aligned}
    &\alpha_{i,j}^{\left( l \right)} = f_{\text{att}} \left( \mathbf{X}_i^{\left(l \right)}, \mathbf{X}_j^{\left(l \right)} \right), \quad \alpha_{i,j}^{\left (l \right )} \in \mathbb{R}^{d}, \\
    &\mathbf{X}_i^{\left (l+1 \right )} = \sum_{j \in \mathcal{N}_i} \left (\alpha_{i,j}^{\left (l \right )}+\mathbf{P}_{i, j}^{\left (l \right )}\right) \odot \text{MLP} \left( \mathbf{X}_j^{\left (l \right)} \right),
\end{aligned}
\end{equation}
where $f_{\text{att}}$ is the attention function to calculate the weights of neighbors and $\odot$ indicates the element-wise multiplication.
In general, there are many MPNNs to implement the attention function. For example, if the encoder is GAT~\cite{velivckovic2018graph}, then $d_h$ corresponds to the number of attention heads, and the attention function is defined as
\begin{equation}
    \alpha_{i,j}^{\left(l \right )} = \text{LeakyReLU} \left( \mathbf{w}^T \left[\mathbf{W}^{\left (l \right )}\mathbf{X}_i^{\left (l \right )} \parallel \mathbf{W}^{\left (l \right )}\mathbf{X}_j^{\left (l \right )} \right] \right), \quad \mathbf{w} \in \mathbb{R}^{2d_h}.
\end{equation}
For GatedGCN, $d_h$ is equal to the dimension of the node representation, and the attention function is defined as
\begin{equation}
    \alpha_{i,j}^{\left(l \right )} = \text{Sigmoid} \left(\mathbf{W}_1^{\left (l \right )}\mathbf{X}_i^{\left (l \right )} + \mathbf{W}_2^{\left (l \right )}\mathbf{X}_j^{\left (l \right )} \right).
\end{equation}
Without loss of generality, we omit edge features in the formulation. Regarding the normalization function in Figure \ref{fig:encoder}, GAT uses the Softmax function, and GatedGCN applies degree normalization.

\vspace{2pt}
\noindent \textbf{Position Path: Refine Node Positions.}
The feature path improves the expressiveness of GraphPAE, but still lacks diverse frequency information.
To solve this issue, the position path uses the learned attention weights to update the position representations
\begin{equation}
    \mathbf{P}_{i,j}^{\left ( l+1\right )} = \alpha_{i, j}^{\left ( l\right )} + \mathbf{P}_{i,j}^{\left ( l\right )}.
\end{equation}
Intuitively, the attention weights gradually refine position representations to approach the ground-truth node positions.
Notably, the relative distance $\mathbf{P}$ is calculated based on the eigenvectors of graph Laplacian, which contains various frequency information, from low-frequency $\mathbf{u}_1$ to high-frequency $\mathbf{u}_K$.
Therefore, leveraging the refined position representations to reconstruct the relative distance can force the encoder to learn diverse frequency information.
We find that this design is quite useful in the heterophilic node classification task, where high-frequency information dominates the classification performance. Experiments can be seen in Section \ref{sec:exp_node_classification}.

\subsection{GraphPAE Decoder}
\label{session:decoder}
So far, we have described the encoder of GraphPAE, which effectively learns both node and position representations.
The decoder then utilizes these representations to recover the corrupted information.
In practice, we find that corrupting both node features and positions simultaneously can negatively impact performance, as the recovery process in one path depends on the information from the other.
To resolve this, we corrupt only one path's data during training, leaving the other path's data intact.

\vspace{2pt}
\noindent \textbf{Feature Reconstruction.} 
The node representations are learned by masking node features and preserving the node positions
\begin{equation}
    \widetilde{\mathbf{X}}_i^{(L)}, \mathbf{P}_i^{(L)} = f_{\text{enc}} \left ( \widetilde{\mathbf{X}}_i^{\left (0 \right )}, \left \{\widetilde{\mathbf{X}}_j^{\left (0 \right )}\right \}_{j \in \mathcal{N}_i}, \mathbf{P}_{i}^{\left(0 \right)} \right),
\end{equation}
To reconstruct the original node features, we apply a feature decoder $f_{\text{dec}}^{X}$ to map the node representations back to the feature space. The reconstruction process is defined as
\begin{equation}
    \mathbf{X}^{'}_i = f_{\text{dec}}^{X} \left( \widetilde{\mathbf{X}}_i^{\left( L \right)} \right),
\end{equation}
where $\mathbf{X}^{'}_i$ is the is the reconstructed feature of node $v_i$. 
We follow GraphMAE and use the scaled cosine error (SCE) as the loss function of feature reconstruction
\begin{equation}
\label{eq:loss_feat}
    \mathcal{L}_{\text{feat}} = \frac{1}{|\widetilde{\mathcal{V}}|} \sum_{v_i \in \widetilde{\mathcal{V}}} \left( 1 - \frac{\mathbf{X}_i^T \mathbf{X}_{i}^{'}}{\|\mathbf{X}_i\| \cdot \|\mathbf{X}_{i}^{'}\|} \right)^\gamma,
\end{equation}
where $\gamma \geq 1$ is a hyperparameter.

\vspace{2pt}
\noindent \textbf{Position Reconstruction.}
The position representations are learned by offsetting eigenvectors and preserving the original node features
\begin{equation}
    \mathbf{X}_i^{(L)}, \widetilde{\mathbf{P}}_i^{(L)} = f_{\text{enc}} \left ( \mathbf{X}_i^{\left (0 \right )}, \left \{\mathbf{X}_j^{\left (0 \right )}\right \}_{j \in \mathcal{N}_i}, \widetilde{\mathbf{P}}_{i}^{\left(0 \right)} \right),
\end{equation}
Similarly, a position decoder $f_{\text{dec}}^{P}$ is used to recover the original pair-wise node distances
\begin{equation}
\label{eq:loss_pos}
    \mathbf{P}_{i,j}^{'} = f_{\text{dec}}^{P} \left( \widetilde{\mathbf{P}}_{i, j}^{\left(L\right)} \right)
\end{equation}
For position reconstruction loss, we adopt Huber loss~\cite{huber1992robust}, which can make smooth gradients for better convergence
\begin{equation}
\begin{aligned}
        &\mathcal{L}_{\text{pos}}^{i,j} =
\begin{cases}
\frac{ \left( \mathbf{P}_{i,j}^{\prime} - \mathbf{P}_{i,j} \right)^2 }{2}, & \text{if } |\mathbf{P}_{i,j}^{\prime} - \mathbf{P}_{i,j}| < 1 \\
|\mathbf{P}_{i,j}^{\prime} - \mathbf{P}_{i,j}| - \frac{1}{2}, & \text{otherwise}
\end{cases} \\
&\mathcal{L}_{\text{pos}} = \frac{1}{\sum_{v_i \in \widetilde{\mathcal{V}}}\left | \mathcal{N}_i \right | }\sum_{v_i \in \widetilde{\mathcal{V}}, j \in \mathcal{N}_i}{\mathcal{L}_{\text{pos}}^{i,j}}
\end{aligned}
\end{equation}
The overall loss function is formulated as a weighted combination of the feature and position reconstruction losses
\begin{equation}
    \mathcal{L} = \mathcal{L}_{\text{feat}} + \alpha \mathcal{L}_{\text{pos}}
    \label{eq:total_loss}
\end{equation}
where $\alpha$ is the hyperparameter. The pseudocode of GraphPAE is presented in Algorithm \ref{algo}.

\begin{algorithm}[t]
\caption{GraphPAE.}
\label{algo}
\begin{algorithmic}[1]
  \State \textbf{Input:} Graph $\mathcal{G} = \{\mathcal{V}, \mathcal{E}, \mathbf{X}\}$, masking ratio $r$, epochs $T$, and noise scale $\mu_p$.

  \State \textbf{Preprocess:} Compute top-$K$ eigenvectors $\mathbf{U}$ and node distance matrix $\mathbf{P}$ with $\mathbf{U}$.

  \State \textbf{Init:} Encoder $f_{\text{enc}}$, feature decoder $f_{\text{dec}}^X$, position decoder $f_{\text{dec}}^P$, and learnable token $\mathbf{x}_{[M]}$.

  \For{$t = 1$ to $T$}

    \State Randomly select $r|\mathcal{V}|$ nodes from $\mathcal{V}$ to form $\widetilde{\mathcal{V}}$.
    \State Replace features in $\widetilde{\mathcal{V}}$ with $\mathbf{x}_{[M]}$ to obtain $\widetilde{\mathbf{X}}$.
    \State Add noise $\delta \sim \mathcal{U}(-\mu_p, \mu_p)$ to eigenvectors of nodes in $\widetilde{\mathcal{V}}$ for $\widetilde{\mathbf{U}}$; compute corrupted distances $\widetilde{\mathbf{P}}$ with $\widetilde{\mathbf{U}}$.

  \Comment{Data Corruption}
  
    \State Encode $\widetilde{\mathbf{X}}$ with $\mathbf{P}$ via $f_{\text{enc}}$ for node representations $\widetilde{\mathbf{X}}^{(L)}$.
    
    \State Encode $\widetilde{\mathbf{P}}$ with $\mathbf{X}$ via $f_{\text{enc}}$ for positional encodings $\widetilde{\mathbf{P}}^{(L)}$.

  \Comment{Encoder}

    \State Decode $\widetilde{\mathbf{X}}^{(L)}$ and $\widetilde{\mathbf{P}}^{(L)}$ via $f_{\text{dec}}^X$ and $f_{\text{dec}}^P$ to reconstruct features $\mathbf{X}'$ and distances $\mathbf{P}'$.
    
    \State Compute $\mathcal{L}_{\text{feat}}$ with $\mathbf{X}$ and $\mathbf{X}'$, and $\mathcal{L}_{\text{pos}}$ with $\mathbf{P}$ and $\mathbf{P}'$.
    \State $\mathcal{L} = \mathcal{L}_{\text{feat}} + \alpha \mathcal{L}_{\text{pos}}$.

  \Comment{Decoder}

    \State Update $f_{\text{enc}}, f_{\text{dec}}^X$, and $f_{\text{dec}}^P$ by minimizing $\mathcal{L}$.
  \EndFor

  \State \Return Trained encoder $f_{\text{enc}}$.
\end{algorithmic}
\end{algorithm}

\section{Experiments}
In this section, we conduct extensive experiments, including node classification, graph prediction, and transfer learning on large-scale molecule graphs to verify the effectiveness of GraphPAE. Moreover, we perform ablation studies on position reconstruction and dual-path design. 
Finally, we analyze the influence of the loss of weight and the number of eigenvectors.

\begin{table*}
\caption{Node classification results of different graph self-supervised learning, mean accuracy (\%) ± standard deviation. Bold indicates the best performance and underline means the runner-up.}
\setlength{\tabcolsep}{10pt}
  \label{tab:expnodeclassification}
  \resizebox{0.9 \linewidth}{!}{
  \begin{tabular}{ccccccc}
\toprule
\multirow{3}{*}{\textbf{Dataset}} & \multicolumn{4}{c}{Small Graphs} & \multicolumn{2}{c}{Large Graphs} \\
\cmidrule(lr){2-5}
\cmidrule(lr){6-7} 
& BlogCatalog & Chameleon & Squirrel & Actor & arXiv-year & Penn94 \\
\midrule
Supervised & 80.52±2.10 & 80.02±0.87 & 71.91±1.03 & 33.93±2.47 &  46.02±0.26 &  81.53±0.55 \\
\midrule
DGI & 72.07±0.16 & 43.83±0.14 & 34.56±0.10 & 27.98±0.09 & -  & - \\
BGRL & 79.74±0.46 & 61.24±1.07 & 43.24±0.52 & 26.61±0.57 & \underline{41.43±0.04} & 63.31±0.49 \\
MVGRL & 63.24±0.94 & 73.19±0.42 & 60.09±0.44 & 34.64±0.20 & - & - \\
CCA-SSG & 74.00±0.28 & 75.00±0.75 & 61.58±1.98 & 27.79±0.58 & 40.78±0.01 & 62.63±0.20 \\
Sp$^2$GCL & 72.73±0.46 & 78.88±1.04 & 62.61±0.87 & \underline{34.70±0.92} & 39.09±0.02 & 68.80±0.45 \\
\midrule
VGAE & 60.47±1.84 & 62.32±1.90 & 42.50±1.35 & 31.57±0.75 & 36.39±0.21 & 55.31±0.28 \\
GraphMAE & 79.90±1.13 & \underline{79.50±0.57} & 61.13±0.60 & 32.15±1.33 & 40.30±0.04 & 67.97±0.21 \\
GraphMAE2 & 77.34±0.12 & 79.13±0.19 & \underline{70.27±0.88} & 34.48±0.26 & 38.97±0.03 & 67.86±0.42 \\
MaskGAE & 73.10±0.08 & 74.50±0.87 & 68.53±0.44 & 33.44±0.34 & 40.59±0.04 & 63.84±0.03  \\
S2GAE & 75.76±0.43 & 60.24±0.37 & 68.60±0.56 & 26.17±0.38 & 40.32±0.12 & \underline{70.24±0.09} \\
AUG-MAE & \underline{82.03±0.69} & 70.10±1.88 & 62.57±0.67 & 33.42±0.38 & 37.10±0.13 & 69.90±0.43 \\
\midrule
GraphPAE & \textbf{85.76±1.22} & \textbf{80.51±1.25} & \textbf{72.05±1.40} & \textbf{38.55±1.35} & \textbf{41.85±0.04} & \textbf{71.79±0.37} \\
\bottomrule
\end{tabular}
}
\end{table*}

\begin{table*}
\caption{Graph regression and classification results of different graph self-supervised learning on OGB datasets. Bold indicates the best performance and underline means the runner-up. $\downarrow$ means lower the better and $\uparrow$ means higher the better.}
  \label{tab:expgraphprediction}
  \setlength{\tabcolsep}{9pt}
  \resizebox{\linewidth}{!}{
  \begin{tabular}{cccccccc}
\toprule
\textbf{Task} & \multicolumn{3}{c}{Regression (Metric: RMSE $\downarrow$)} & \multicolumn{3}{c}{Classification (Metric: ROC-AUC\% $\uparrow$)} \\
\cmidrule(lr){2-4} 
\cmidrule(lr){5-8}
\textbf{Dataset} & molesol & molipo & molfreesolv & molbace & molbbbp & molclintox & moltocx21 \\
\midrule
Supervised & 1.173±0.057 & 0.757±0.018 & 2.755±0.349 & 80.42±0.96 & 68.17±1.48 & 88.14±2.51 & 74.91±0.51 \\
\midrule
InfoGraph & 1.344±0.178 & 1.005±0.023 & 10.005±8.147 & 73.64±3.64 & 66.33±2.79 & 64.50±5.32 & 69.74±0.57 \\
GraphCL & 1.272±0.089 & 0.910±0.016 & 7.679±2.748 & 73.32±2.70 & 68.22±2.19 & 74.92±4.42 & 72.40±1.07  \\
MVGRL & 1.433±0.145 & 0.962±0.036 & 9.024±1.982 & 74.88±1.43 & 67.24±3.19 & 73.84±2.75 & 70.48±0.83 \\
JOAO & 1.285±0.121 & 0.865±0.032 & 5.131±0.782 & 74.43±1.94 & 67.62±1.29 & 71.28±4.12 & 71.38±0.92 \\
Sp$^2$GCL & 1.235±0.119 & \underline{0.835±0.026} & 4.144±0.573 & 78.76±1.43 & \textbf{68.72±1.53} & 80.88±3.86 & 73.06±0.75 \\
\midrule
GraphMAE & \underline{1.050±0.034} & 0.850±0.022 & 2.740±0.233 & 79.14±1.31 & 66.55±1.78 & 80.56±5.55 & 73.84±0.58  \\
GraphMAE2 & 1.225±0.081 & 0.885±0.019 & 2.913±0.293 & \underline{80.74±1.53} & 67.67±1.44 & 75.75±3.65 & 72.93±0.69 \\
StructMAE & 1.499±0.043 & 1.089±0.002 & 2.568±0.262 & 77.75±0.42 & 65.66±1.16 & 79.42±4.56 & 71.13±0.61 \\
AUG-MAE & 1.248±0.026 & 0.917±0.013 & \underline{2.395±0.158} & 78.54±2.49 & 67.05±0.63 & \underline{82.66±1.98} & \underline{74.33±0.07} \\
\midrule
GraphPAE & \textbf{1.015±0.045} & \textbf{0.810±0.018} & \textbf{2.058±0.188} & \textbf{81.11±1.24} & \underline{68.56±0.71} & \textbf{82.69±3.39} & \textbf{74.46±0.54} \\
\bottomrule
\end{tabular}
  }
\end{table*}

\subsection{Node Classification} \label{sec:exp_node_classification}

\noindent \textbf{Dataset.} We evaluate the performance of GraphPAE on 6 representative heterophilic graphs: BlogCatalog~\cite{meng2019co}, Chameleon, Squirrel, Actor~\cite{pei2020geom}, arXiv-year~\cite{hu2020open}, and Penn94~\cite{traud2012social}. Specifically, arXiv-year and Penn94 are large-scale graphs (> 40,000) to evaluate the scalability of the methods. 
As these datasets place greater emphasis on high-frequency information, they are well suited to evaluate GraphPAE’s ability to capture diverse frequency components.

\vspace{2pt}
\noindent \textbf{Baselines and Settings.} We benchmark GraphPAE against a wide range of graph self-supervised baselines, which can be roughly divided into: \textbf{(1) contrastive learning,} \emph{i.e.}, DGI~\cite{velivckovic2018deep}, BGRL~\cite{thakoor2021bootstrapped}, MVGRL~\cite{hassani2020contrastive}, CCA-SSG~\cite{zhang2021canonical}, and $\text{Sp}^2\text{GCL}$~\cite{bo2024graph}. \textbf{(2) graph autoencoders,} \emph{i.e.}, VGAE~\cite{kipf2016variational}, GraphMAE~\cite{hou2022graphmae}, GraphMAE2~\cite{hou2023graphmae2}, MaskGAE~\cite{li2023s}, S2GAE~\cite{tan2023s2gae}, and AUG-MAE~\cite{wang2024rethinking}. We use GAT with 4 heads and 1024 hidden units as the encoder for all methods, and the number of layers is searched in the range of \{2, 3\}. Moreover, we adopt two-layer MLPs with ReLU activation as both feature and position decoders for efficiency and scalability.
In the evaluation protocol, we freeze the encoder and generate node representations. The node representations are then input into a linear classifier for training with labeled data and inference for node classification. For all methods, we use the Adam optimizer and run 10 times on each graph.

\vspace{2pt}
\noindent \textbf{Results.} As shown in Table \ref{tab:expnodeclassification}, GraphPAE consistently outperforms all baseline methods across all datasets, demonstrating the effectiveness of our framework. We highlight two important observations as follows. (1) GraphPAE surpasses both spatial- and spectral-based contrastive learning methods on most datasets, indicating its ability to effectively encode both spatial and spectral patterns through the joint reconstruction of features and positions. (2) GraphPAE also consistently outperforms existing GAEs, including those focused on feature reconstruction, structure reconstruction, and hybrid strategies. This superior performance is mainly attributed to two key factors. First, the incorporation of positional encodings enhances the expressivity of node representations. Second, reconstructing positions encourages the GAE to capture diverse frequency information, leading to high-quality graph representations.

\subsection{Graph Prediction}

\noindent \textbf{Datasets.} For graph-level tasks, we evaluate GraphPAE on 7 OGB datasets~\cite{hu2020open}, including 3 graph regression tasks and 4 graph classification tasks. For all datasets, we use public splits for a fair comparison. We use MSE and ROC-AUC as the evaluation metrics for regression and classification, respectively.

\vspace{2pt}
\noindent \textbf{Baselines and Settings.} We select 5 \textbf{contrastive learning} methods, \emph{i.e.}, InfoGraph~\cite{sun2019infograph}, GraphCL~\cite{you2020graph}, MVGRL~\cite{hassani2020contrastive}, JOAO~\cite{you2021graph}, and $\text{Sp}^2\text{GCL}$~\cite{bo2024graph}, and 4 \textbf{graph autocoders}, \emph{i.e.}, GraphMAE~\cite{hou2022graphmae}, GraphMAE2~\cite{hou2023graphmae2}, StructMAE~\cite{10.24963/ijcai.2024/241}, and AUG-MAE~\cite{wang2024rethinking}. We use a two-layer GatedGCN as the encoder and set the hidden dimension $d_h=300$. We adopt two-layer MLPs as feature and position decoders. For evaluation, we freeze the encoder to output node representations and input them into pooling functions for graph representations. Similarly, we input the graph representations into a linear classifier to evaluate the performance for downstream tasks. We use Adam optimizer and report the metrics with mean results and standard deviation of 10 seeds.

\vspace{2pt}
\noindent \textbf{Results.} Table \ref{tab:expgraphprediction} summarizes the results of graph-level tasks. GraphPAE consistently achieves superior performance across both regression and classification benchmarks, highlighting its ability to learn high-quality graph representations. Notably, GraphPAE shows competitive performance against contrastive learning methods and even exhibits notable improvements compared against the spectral-based method $\text{Sp}^2\text{GCL}$ on molfreesolv, molbace, and molclintox. In addition, GraphPAE also outperforms other GAEs across most datasets. We attribute these gains to the proposed position corruption-reconstruction strategy, which improves the encoder's ability to identify crucial substructures for downstream tasks.

\begin{table*}
\caption{Quantum chemistry property results of transfer learning on QM9. The best and runner-up results are highlighted with bold and underline, respectively.}
  \label{tab:qm9}
  \resizebox{\linewidth}{!}{
  \begin{tabular}{ccccccccccccc}
\toprule
\textbf{Target} & $\mu$ & $\alpha$ & $\epsilon_{\text{homo}}$ & $\epsilon_{\text{lumo}}$ & $\Delta_{\epsilon}$ & $R^2$ & ZPVE & $U_0$ & $U$ & $H$ & $G$ & $C_v$ \\
\cmidrule(lr){2-13} 
\textbf{Unit} & D & $a_0^3$ & $10^{-2}$meV & $10^{-2}$meV & $10^{-2}$meV & $a_0^2$ & $10^{-2}$meV & meV & meV & meV & meV & cal/mol/K \\
\midrule
GraphCL & 1.035 & 2.321 & 2.030 & \underline{3.667} & 4.523 & 40.725 & \underline{2.063} & 2.461 & \underline{1.745} & \underline{1.734} & \underline{1.751} & \underline{1.747}   \\
GraphMAE & \underline{1.030} & 2.924 & 2.407 & 6.373 & 4.813 & 41.955 & 4.623 & \underline{1.411} & 2.207 & 2.208 & 2.207 & 2.200 \\
Mole-BERT & 1.031 & \underline{1.918} & \underline{1.477} & 4.127 & 4.240 & 44.374 & 2.190 & 2.532 & 2.509 & 2.511 & 2.516 & 2.508 \\
SimSGT & 1.064 & 2.413 & 2.837 & 4.227 & \underline{4.107} & \underline{40.504} & 2.127 & 1.948 & 2.420 & 2.416 & 2.416 & 2.410 \\
\midrule
GraphPAE & \textbf{0.703} & \textbf{0.879} & \textbf{1.199} & \textbf{2.141} & \textbf{2.289} & \textbf{36.480} & \textbf{0.502} & \textbf{0.510} & \textbf{0.639} & \textbf{0.639} & \textbf{0.641} & \textbf{0.643} \\
\bottomrule
\end{tabular}
}
\end{table*}

\begin{table*}
\caption{Ablation studies of position reconstruction and framework design on node- and graph-level tasks. \emph{Exp No.}: the number of different experimental settings. \emph{Corrupt Info.}: corrupted information during training. \emph{Recon Info.}: information to be reconstructed during training. \textbf{Bold} indicates the best performance.}
  \label{tab:ablation}
  \setlength{\tabcolsep}{8pt}
  \resizebox{\linewidth}{!}{
  \begin{tabular}{cccccc|ccccc}
\toprule
Exp & \multicolumn{2}{c}{Corrupt Info.} & \multicolumn{2}{c}{Recon Info.} & \multirow{2}{*}{Dual-Path} & \multicolumn{2}{c}{Node-level} & \multicolumn{3}{c}{Graph-level} \\
\cmidrule(lr){2-3} \cmidrule(lr){4-5} \cmidrule(lr){7-8} \cmidrule(lr){9-11}
No. & Feature & Position & Feature & Position &  & Blog ($\uparrow$) & Squirrel ($\uparrow$) & Bace ($\uparrow$) & Bbbp ($\uparrow$)  & Freesolv ($\downarrow$) \\
\midrule
a & \Checkmark & \Checkmark & \Checkmark &  & & 82.8±1.7 & 66.4±1.6 & 78.4±1.2 & 66.4±1.7 & 2.79±0.40 \\
b & \Checkmark & \Checkmark & \Checkmark & \Checkmark & & 83.5±1.0 & 68.5±0.9 & 78.9±2.1 & 66.8±0.6 & 2.44±0.36 \\
c & \Checkmark &  & \Checkmark &  &  & 84.6±1.6 & 71.3±0.9 & 79.4±3.4 & 67.7±0.9 & 2.20±0.14 \\
d & \Checkmark & \Checkmark & \Checkmark & \Checkmark & \Checkmark & \textbf{85.8±1.2} & \textbf{72.1±1.4} & \textbf{81.1±1.2} & \textbf{68.6±0.7} & \textbf{2.06±0.19} \\
\bottomrule
\end{tabular}
  }
\end{table*}

\subsection{Transfer Learning}
\noindent \textbf{Settings.} We conduct transfer learning experiments on molecule property prediction to evaluate the generalization ability of GraphPAE. Specifically, we follow the setting of~\cite{liu2024rethinking}, which first pre-trains the encoder on 2 million molecules sampled from ZINC15~\cite{sterling2015zinc}, and then fine-tunes on QM9~\cite{wu2018moleculenet} to predict the quantum chemistry properties. We compare GraphPAE with the state-of-the-art molecule graph pre-training model, \emph{i.e.}, GraphCL~\cite{you2020graph}, GraphMAE~\cite{hou2022graphmae}, Mole-BERT~\cite{xia2023mole}, and SimSGT~\cite{liu2024rethinking}. We use a five-layer GatedGCN with 300 hidden units. Upon finishing pre-training, a two-layer MLP is attached after the graph representations for property prediction. During the fine-tuning protocol, the encoder and attached MLP are trained together on labeled data of downstream tasks. Consistent with~\cite{liu2024rethinking}, we divided QM9 into train/validation/test sets with 80\%/10\%/10\% by scaffold splits. We use Adam optimizer and run it 5 times to report average MAE and standard deviation. 

\vspace{2pt}
\noindent \textbf{Results.} 
From Table \ref{tab:qm9}, we observe that GraphPAE demonstrates robust generalization across all prediction targets compared to state-of-the-art models. 
Among the baselines, Mole-BERT and SimSGT are specially designed for molecular graph pretraining, incorporating customized reconstruction objectives tailored to the characteristics of molecular graphs.
Specifically, Mole-BERT pretrains a discrete codebook within the subgraph of nodes, inherently capturing both features and local structural patterns. During reconstruction, the model replaces the raw features with codebook entries as predictive targets.
SimSGT employs a tokenizer to encode feature and local structure information, using tokenized output as reconstruction targets. 
Despite these tailored strategies, GraphPAE consistently outperforms these models. The superiority is largely attributed to the effective positional encodings obtained by position corruption-reconstruction, enabling the graph representations to capture crucial substructures for prediction targets such as cycles.

\subsection{Ablation Studies}

To verify the effectiveness of important designs of GraphPAE, we conduct extensive ablation experiments. We select 2 datasets for node-level tasks and 3 datasets for graph-level tasks.

\vspace{2pt}
\noindent \textbf{Effectiveness of Position Reconstruction.} To achieve position corruption-reconstruction, GraphPAE inevitably introduces positional encodings into the encoder. 
Firstly, we verify that the superior performance is attributed not only to the introduction of positional encodings but also to the position reconstruction. 
We conducted two sets of comparative experiments in Table \ref{tab:ablation}: one comparing Exp a with Exp b, and another comparing Exp c with Exp d. 
We have the following summaries: 
(1) In Exp a, both features and positions are corrupted, but only the features are reconstructed. In Exp b, both features and positions are reconstructed. We observe that Exp b performs better than Exp a, demonstrating the effectiveness of position reconstruction. 
(2) In Exp c, we remove both position corruption and reconstruction from GraphPAE. Comparing Exp d against Exp c, we find that the encoder incorporating position reconstruction consistently improves performance compared to feature reconstruction alone.

\vspace{2pt}
\noindent \textbf{Effectiveness of Dual-Path Reconstruction.} 
Comparing Exp b and Exp c in Table \ref{tab:ablation}, we observe that position reconstruction without the dual-path architecture fails to improve performance and may even lead to degradation. 
Specifically, in Exp b, both masked features and noisy positions are reconstructed directly from node representations. However, recovering relative distances from node representations generated with corrupted positions is more difficult than recovering them from refined distance encoding directly. 
Rather than enhancing node representations or capturing diverse frequency information, the corrupted positional inputs introduce additional noise during training. 
As a result, applying position corruption-reconstruction without appropriate design hinders the model’s ability to learn meaningful structural patterns.
Based on these observations, we propose a dual-path architecture that disentangles feature and relative distance encodings for reconstruction. Experimental results from Exp d demonstrate that this strategy consistently achieves the best performance across datasets, validating the effectiveness of our dual-path encoder design.

\vspace{2pt}
\noindent \textbf{Improvement Attributed to $\mathcal{L}_{pos}$ and PEs.} 
Since integrating positional encodings also enhances representation learning, it remains unclear how much of the performance gain in GraphPAE stems from the position corruption-reconstruction mechanism, and how much from the positional information itself.
Therefore, we conduct ablation studies on both $\mathcal{L}_{pos}$ and positional encodings. Specifically, we remove $\mathcal{L}_{pos}$ from Equation \ref{eq:total_loss} while keeping PEs, to evaluate the effect of position reconstruction.
Then, we remove both $\mathcal{L}_{pos}$ and PEs to isolate the impact of positional information. 
The results are reported in Table \ref{tab:ablation_le_pe}. 
Notably, even with positional encodings retained, removing $\mathcal{L}_{pos}$ consistently leads to a noticeable performance decline across all datasets, highlighting the importance of the proposed position corruption-reconstruction mechanism in enhancing GAEs.

\begin{table}
\caption{Ablation studies of $\mathcal{L}_{pos}$ and positional encodings.}
  \label{tab:ablation_le_pe}
  \setlength{\tabcolsep}{8pt}
  \resizebox{\linewidth}{!}{
  \begin{tabular}{cccc}
\toprule
\textbf{Methods} & Bace ($\uparrow$) & Bbbp ($\uparrow$)  & Freesolv ($\downarrow$) \\
\midrule
GraphPAE & \textbf{81.11±1.24} & \textbf{68.56±0.71} & \textbf{2.058±0.19} \\
w/o $\mathcal{L}_{pos}$ & 79.40±3.45 & 67.74±0.92 & 2.204±0.14 \\
w/o $\mathcal{L}_{pos}$ \& PEs & 79.14±1.31 & 66.55±1.78 & 2.740±0.23 \\
\bottomrule
\end{tabular}
  }
\end{table}

\begin{figure}[t]
    \subfigure[Molbbbp]{
        \label{fig:hyperparameter_alpha_bbbp}
        \centering
        \includegraphics[width=0.46\linewidth]{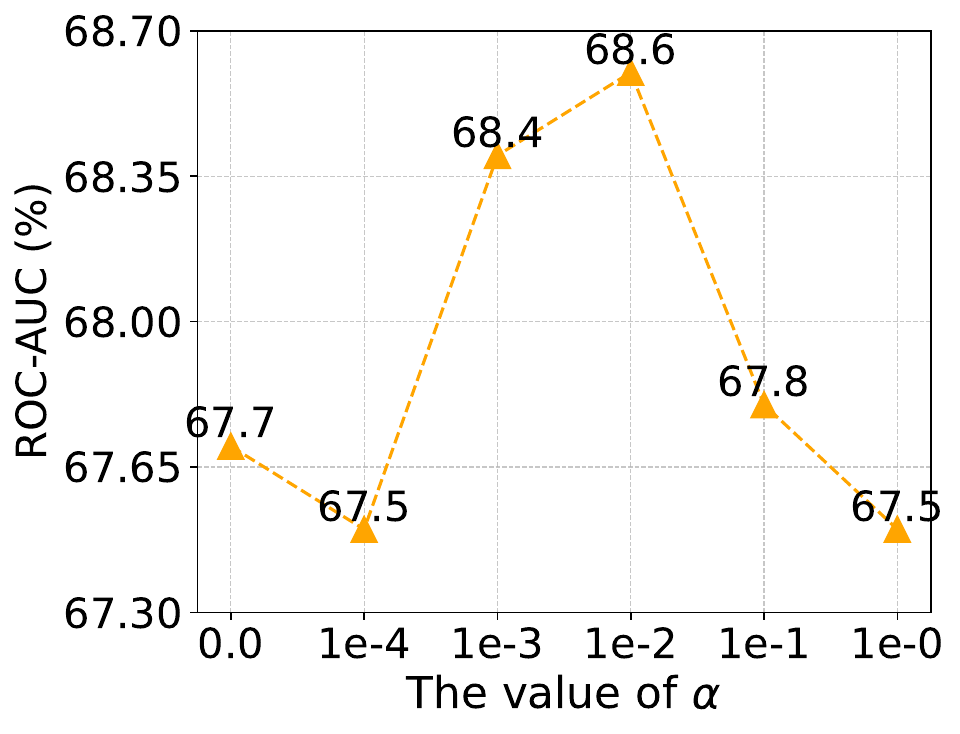}
    }
    \subfigure[BlogCatalog]{
        \label{fig:hyperparameter_alpha_blog}
        \centering
        \includegraphics[width=0.46\linewidth]{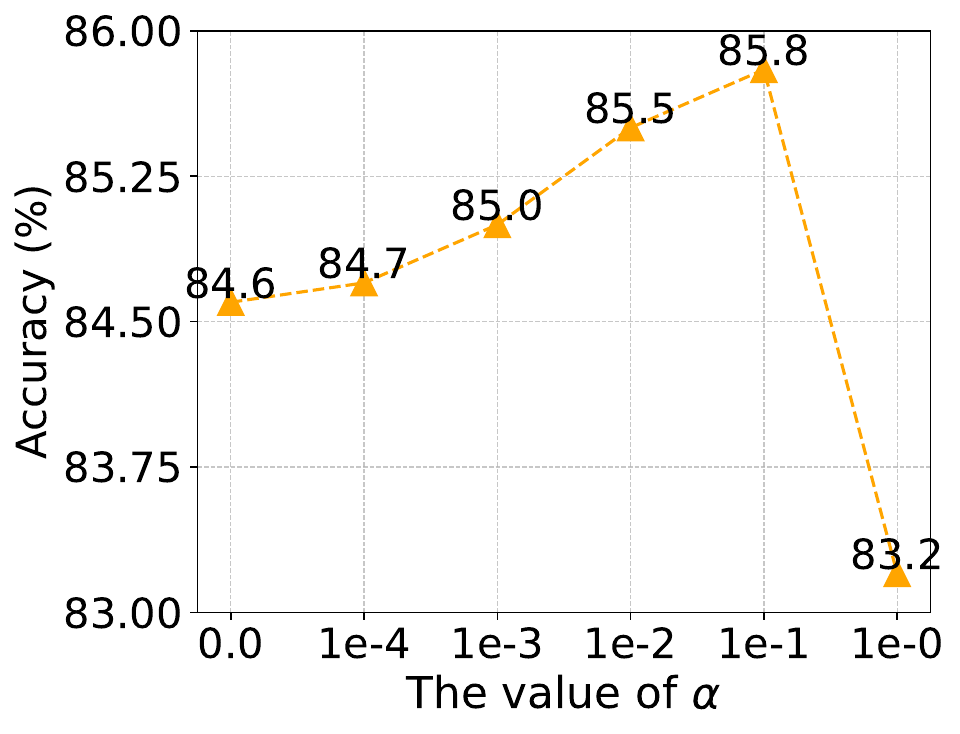}
    }
    \subfigure[Squirrel]{
        \label{fig:hyperparameter_k_squirrel}
        \centering
        \includegraphics[width=0.46\linewidth]{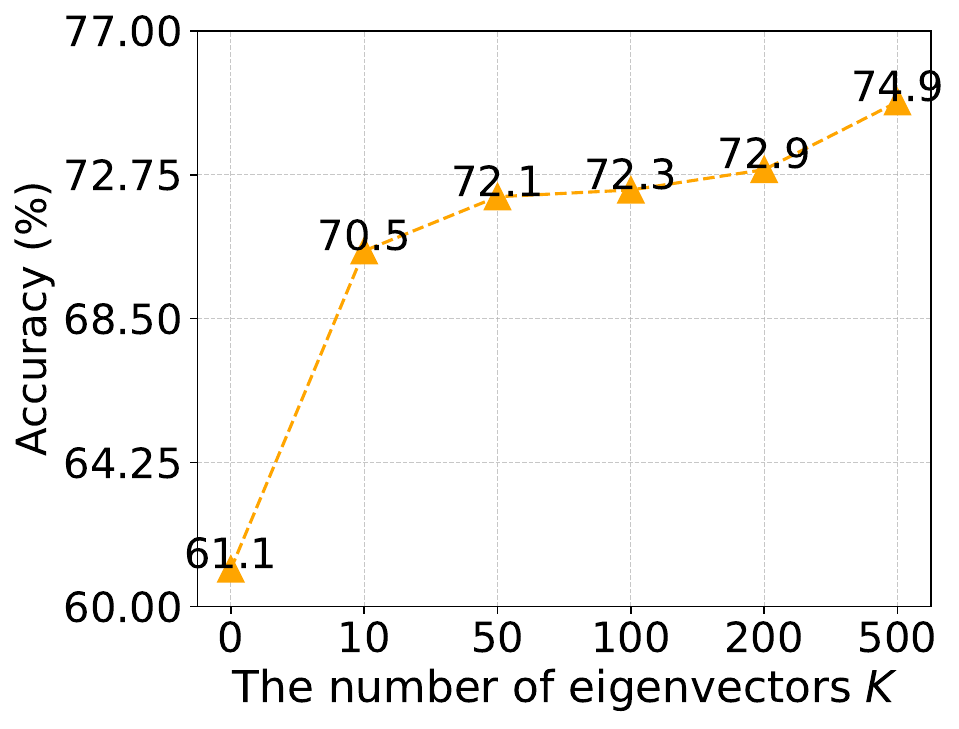}
    }
    \subfigure[Chameleon]{
        \label{fig:hyperparameter_k_chameleon}
        \centering
        \includegraphics[width=0.46\linewidth]{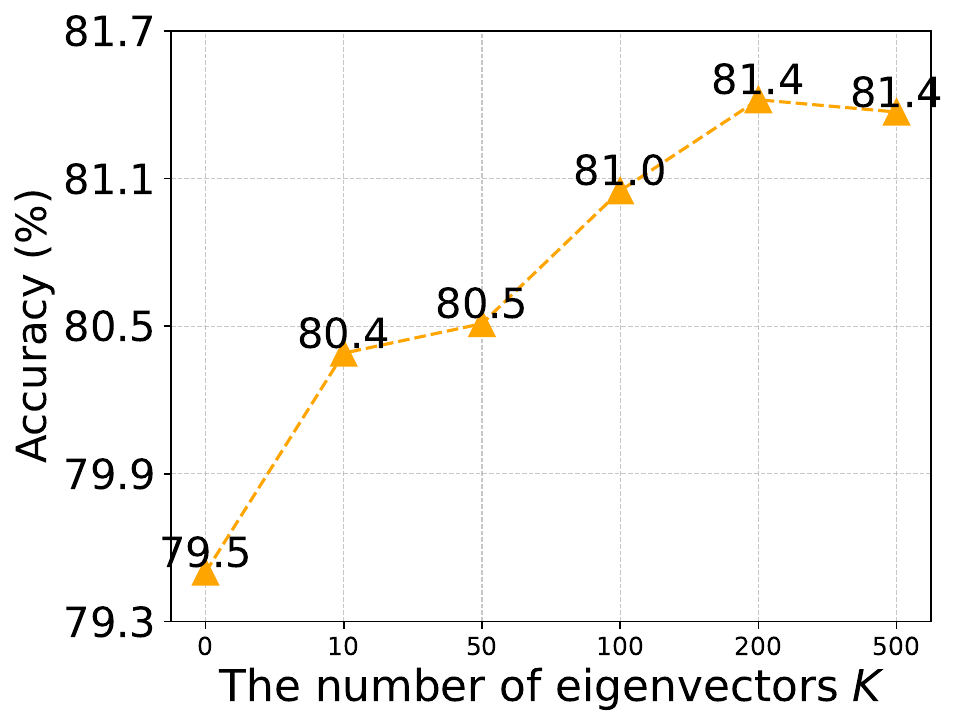}
    }
    \caption{Influence of the loss weight $\alpha$ and the number of eigenvectors $K$.}
    \label{fig:hyperparameter}
\end{figure}

\subsection{Parameters Analysis}
We conduct additional parameter analysis of the loss weight $\alpha$ and the number of eigenvectors $K$.

\vspace{2pt}
\noindent \textbf{Influence of the loss weight $\alpha$ in GraphPAE.} Figure \ref{fig:hyperparameter_alpha_bbbp} and \ref{fig:hyperparameter_alpha_blog} present the hyperparameter analysis of $\alpha$ to further examine the influence of the positional reconstruction loss $\mathcal{L}_{pos}$. 
We summarize the key observations as follows:
(1) As $\alpha$ increases within a certain threshold, the performance improves progressively, indicating that position reconstruction contributes positively to the quality of learned representations.
(2) However, excessively large values of $\alpha$ lead to performance degeneration. This suggests assigning a high weight to position reconstruction may cause the encoder to overemphasize positional information at the expense of overall representation learning.
(3) While the optimal value of $\alpha$ varies slightly across datasets, we observe that the best-performing range typically lies within \{1e-3, 1e-2, 1e-1\}, making it easy to search in practice. Additional hyperparameter details are provided in Appendix \ref{app:hyperparameters}.

\vspace{2pt}
\noindent \textbf{Influence of the number of eigenvectors $K$ in GraphPAE.} We further analyze the effect of the number of eigenvectors $K$ on performance using the Squirrel and Chameleon datasets, as shown in Figure \ref{fig:hyperparameter_k_squirrel} and \ref{fig:hyperparameter_k_chameleon}. As $K$ increases, performance generally improves due to the model's ability to capture a wider range of frequency information. However, once $K$ surpasses a certain threshold, the performance gains become marginal. Therefore, to balance effectiveness and computational efficiency, we adopt a moderate value of $K$, typically in the range of [50, 100], which provides a good trade-off in practice.

\section{Conclusion}
In this paper, we propose GraphPAE, a graph autoencoder that reconstructs both node features and relative distance. To enhance models' expressivity to distinguish intricate patterns and the ability to integrate various frequency information, GraphPAE introduces position corruption and recovery into GAEs and designs a dual-path reconstruction strategy. Extensive experiments, including node classification, graph prediction, and transfer learning, demonstrate the superiority of our GraphPAE.

\begin{acks}
This work is supported in part by the National Natural Science Foundation of China (No. 62192784, U22B2038, 62472329).
\end{acks}

\bibliographystyle{ACM-Reference-Format}
\bibliography{sample-base}

\appendix


\section*{Appendices}

\section{Spectral Analysis of Masking Strategies}
\label{app:freq}

\noindent \textbf{Masking Strategies for Node Features and Edges.} We conduct the experiments in the Squirrel datasets with a 20\% masking ratio. For node feature masking, given the node feature matrix $\mathbf{X}$, we randomly mask 20\% of the node in the dataset and set the feature vector to zero. The remaining 80\% of nodes retain their original feature. The corrupted feature matrix is defined as $\widetilde{\mathbf{X}}$. For edge masking, given the original graph structure $\mathbf{A}$, we randomly remove 20\% of the edges from the graph while preserving the node features. This process results in a structurally corrupted graph with a new adjacency matrix $\widetilde{\mathbf{A}}$.

\vspace{2pt}
\noindent \textbf{Computation of Frequency Magnitudes.} 
First, we calculate Laplacian matrix by:
\begin{equation}
    \mathbf{L} = \mathbf{I}_{N} - \mathbf{D}^{-1/2} \mathbf{A} \mathbf{D}^{-1/2}
    \label{eq:lap}
\end{equation}
The eigenvectors of $\mathbf{L}$ are decomposed as $\mathbf{L} = \mathbf{U} \bm{\Lambda} \mathbf{U}^{\top}$, where $\bm{\Lambda}=\text{diag}(\{\lambda_{i}\}_{i=1}^{N})$, $\mathbf{U}=[\mathbf{u}_{1}, \cdots, \mathbf{u}_{N}] \in \mathbb{R}^{N \times N}$, and the eigenvalues are ordered as $0 \leq \lambda_{1} \leq \cdots \leq \lambda_{N} \leq 2$. $\lambda_{i}$ reflects the frequency magnitude of corresponding eigenvector $\mathbf{u}_{i}$ over the graph.
We transform node features into the spectral domain by
\begin{equation}
    \mathbf{X}^s = \mathbf{U}^\top \mathbf{X}, \quad \mathbf{X}^s \in \mathbb{R}^{N \times d}.
    \label{eq:a_transform}
\end{equation}
In particular, the i-th row of $\mathbf{X}^s$ corresponds to the frequency magnitude at eigenvalue $\lambda_i$. With $d$-dimensional $\mathbf{X}^s$, we first calculate the mean magnitude by:
\begin{equation}
    \bar{\mathbf{X}}^s = \frac{1}{d}\sum_{i=1}^{d} \mathbf{X}_{:,i}^s, \quad \bar{\mathbf{X}}^s\in \mathbb{R}^N
\end{equation}
For a given frequency band $[f_1, f_2]$, we extract the set of indices $\mathcal{I}$ corresponding to eigenvalues within this range:

\begin{equation}
    \mathcal{I} = \{ i \mid f_1 \leq \lambda_i \leq f_2 \}
\end{equation}
The average frequency magnitude within this band is computed as:
\begin{equation}
    \bar{\mathbf{X}}^s_{[f_1, f_2]} = \frac{1}{|\mathcal{I}|} \sum_{i \in \mathcal{I}} \bar{\mathbf{X}}^s[i], \quad \bar{\mathbf{X}}^s_{[f_1, f_2]} \in \mathbb{R}
    \label{eq:a_freq}
\end{equation}
For feature masking case, we calculate $\mathbf{L}$ by substituting $\mathbf{A}$ into Equation \ref{eq:lap}. Then we decompose $\mathbf{L}$ and get $\mathbf{U}$. By substituting $\mathbf{U}$, $\mathbf{X}$, and $\widetilde{\mathbf{X}}$ in Equation (\ref{eq:a_transform})-(\ref{eq:a_freq}), we can get the frequency magnitude of original and corrupted graphs. The comparison of differences is shown in Figure \ref{fig:mask_feat}. For edge masking case, we calculate $\mathbf{L}$ and $\widetilde{\mathbf{L}}$ by substituting $\mathbf{A}$ and $\widetilde{\mathbf{A}}$ into Equation \ref{eq:lap}. Then we decompose $\mathbf{L}$ and $\widetilde{\mathbf{L}}$, and get $\mathbf{U}$ and $\widetilde{\mathbf{U}}$. By substituting $\mathbf{U}$, $\widetilde{\mathbf{U}}$, and $\mathbf{X}$ in Equation (\ref{eq:a_transform}) - (\ref{eq:a_freq}), we can also get the frequency magnitude of original and corrupted graphs as shown in Figure \ref{fig:mask_edge}.

\begin{table}
\caption{Statistics of node classification datasets.}
\label{tab:a_node_sta}
\setlength{\tabcolsep}{2pt}
  \resizebox{\linewidth}{!}{
\begin{tabular}{lccccl}
\toprule
\textbf{Datasets} & \textbf{Nodes} & \textbf{Edges} & \textbf{Features} & \textbf{Classes} & \textbf{Train/Valid/Test}\\ 
\midrule
BlogCatalog & 5,196 & 343,486 & 8,189 & 6 & 120/1,000/1,000 \\ 
Chameleon & 2,277 & 62,792 & 2,325 & 5 & 1,092/729/456 \\ 
Squirrel & 5,201 & 396,846 & 2,089 & 5 & 2,496/1,664/1,041 \\ 
Actor & 7,600 & 53,411 & 932 & 5 & 3,648/2,432/1,520  \\ 
arXiv-year & 169,343 & 2,315,598 & 128 & 5 & 84671/42335/42337  \\ 
Penn94 & 41,554 & 2,724,458 & 4,814 & 2 & 19,407/9,703/9,705 \\
\bottomrule
\end{tabular}
}
\end{table}

\begin{table}
\caption{Statistics of graph prediction datasets.}
\label{tab:a_graph_sta}
\setlength{\tabcolsep}{1pt}
  \resizebox{\linewidth}{!}{
\begin{tabular}{lccccll}
\toprule
 \textbf{Dataset} & \textbf{Graphs} & \textbf{Avg. Nodes} & \textbf{Avg. Edges} & \textbf{Classes} & \textbf{Task} & \textbf{Metric} \\ 
\midrule
molesol & 1,128 & 13.3 & 13.7 & 1 & Regression & RMSE \\
mollipo & 4,200 & 27.0 & 29.5 & 1 & Regression & RMSE \\
molfreesolv & 642 & 8.7 & 8.4 & 1 & Regression & RMSE \\
molbace & 1,513 & 34.1 & 36.9 & 1 & Binary Class. & ROC-AUC \\
molbbbp & 2,039 & 24.1 & 26.0 & 1 & Binary Class. & ROC-AUC \\
molcintox & 1,477 & 26.2 & 27.9 & 2 & Binary Class. & ROC-AUC \\
moltox21 & 7,831 & 18.6 & 19.3 & 12 & Binary Class. & ROC-AUC \\
\midrule
ZINC15 & 2,000,000 & 26.62 & 57.72 & - & Pre-Training & - \\
QM9 & 1,177,631 & 8.80 & 18.81 & 12 Targets &  Finetuning & MAE \\
\bottomrule
\end{tabular}
}
\end{table}

\section{Experimental Details}
\subsection{Statistics of Datasets}
\label{app:statistics}
We benchmark GraphPAE across various tasks, including node classification, graph prediction, and transfer learning. Specifically, we conduct experiments on six node classification datasets and seven graph-level prediction datasets and select ZINC15 and QM9 for transfer learning.
For datasets in node classification and graph prediction, we utilize public data splits. In transfer learning, we follow~\cite{liu2024rethinking} and divide QM9 into train/valid/test sets with 80\%/10\%/10\% by scaffold split.
Detailed statistics of the node- and graph-level datasets are presented in Table \ref{tab:a_node_sta} and Table \ref{tab:a_graph_sta}, respectively.

\subsection{Hyperparameters}
\label{app:hyperparameters}
Hyperparameter details of node classification and graph prediction are reported in Table \ref{tab:a_node_hype} and Table \ref{tab:a_graph_hype}. 
$r$ denotes the mask ratio and $\alpha$ is the loss weight of $\mathcal{L}_{pos}$. dp and $\text{dp}_{edge}$ represent the dropout of node features and edge features, respectively. 
For transfer learning, we pre-train GraphPAE on 2 million molecules sampled from ZINC15 with 100 epochs. The mask ratio is set to 0.35, and the loss weight for $\mathcal{L}_{pos}$ is 0.01. Both node and edge dropout are set to 0.0, and the number of eigenvectors for positional encoding is 6.
During fine-tuning, the encoder and the attached MLP are jointly trained on the QM9 dataset using an initial learning rate of 0.001 and no weight decay. We apply a dropout rate of 0.1 to both node features and edge features within GNN layers.

\subsection{Additional Experiment Results.}
We further evaluate GraphPAE on homophilic graphs, including Facebook~\cite{rozemberczki2021multi} and Wiki~\cite{mernyei2020wiki}, as well as a large-scale graph ogbn-products~\cite{hu2020open}. Notably, ogbn-products contains 2,449,029 nodes and 61,859,140 edges, making it a suitable benchmark to demonstrate the scalability of GraphPAE.
The experimental results are summarized in Table \ref{tab:additional_exp}, and we make the following observations: 
(1) GraphPAE consistently achieves competitive performance across these different benchmarks.
(2) The performance gains on homophilic graphs are less pronounced compared to those on heterophilic graphs. We conjecture this is because, in homophilic graphs, a small portion of very low-frequency information is sufficient.

\section{Complexity Analysis}
 The eigen-decomposition complexity is $\mathcal{O}(N^3)$. However, we only need to decompose the smallest $K$ eigenvalues, which reduces complexity to $\mathcal{O} \left ( N^2K \right )$. Additionally, the decomposition is performed once per graph, so its cost is amortized over the entire experiment.

During training, GraphPAE introduces an additional $\mathcal{O\left(\mathcal{E} \right)}$ complexity per layer for PE computation and an extra cost for relative node distance reconstruction. Since we only reconstruct corrupted node pairs, the complexity remains much lower than $\mathcal{O}\left(\mathcal{E} \right)$. 
However, this slight overhead enables the encoder to capture more diverse frequency information.

\begin{table}
  \caption{Hyperparameters of node classification.}
  \label{tab:a_node_hype}
  \setlength{\tabcolsep}{7pt}
    \resizebox{\linewidth}{!}{
  \begin{tabular}{lccccccc}
    \toprule
    Dataset & lr & wd & $r$ & $\alpha$ & dp & $\text{dp}_\text{edge}$ & $K$ \\
    \midrule
    BlogCatalog & 0.001 & 0.0 & 0.25 & 0.1 & 0.6 & 0.0 & 50 \\
    Chameleon & 0.001 & 0.0 & 0.25 & 0.01 & 0.0 & 0.0 & 50 \\
    Squirrel & 0.001 & 0.0 & 0.5 & 0.001 & 0.6 & 0.0 & 50 \\
    Actor & 0.0005 & 0.0 & 0.25 & 0.01 & 0.0 & 0.0 & 50 \\
    arXiv-year & 0.001 & 0.0 & 0.5 & 0.01 & 0.0 & 0.0 & 100 \\
    Penn94 & 0.001 & 0.0 & 0.25 & 0.001 & 0.0 & 0.0 & 200 \\
    \bottomrule
  \end{tabular}
  }
\end{table}

\begin{table}
  \caption{Hyperparameters of graph prediction.}
  \label{tab:a_graph_hype}
  \setlength{\tabcolsep}{4pt}
    \resizebox{\linewidth}{!}{
  \begin{tabular}{lccccccccc}
    \toprule
    Dataset & pooling & epoch & lr & wd & $r$ & $\alpha$ & dp & $\text{dp}_\text{edge}$ & $K$ \\
    \midrule
    molesol & sum & 20 & 0.0005 & 0.0 & 0.75 & 0.1 & 0.6 & 0.5 & 8 \\
    molipo & sum & 20 & 0.0005 & 0.0001 & 0.25 & 0.001 & 0.6 & 0.0 & 30 \\
    molfreesolv & sum & 100 & 0.0001 & 0.0 & 0.5 & 0.1 & 0.5 & 0.5 & 15 \\
    molbace & mean & 100 & 0.001 & 0.0 & 0.75 & 0.1 & 0.5 & 0.5 & 30 \\
    molbbbp & mean & 20 & 0.001 & 0.0 & 0.5 & 0.01 & 0.6 & 0.6 & 30 \\
    molclintox & mean & 20 & 0.0001 & 0.0 & 0.25 & 0.01 & 0.6 & 0.0 & 30 \\
    moltocx21 & mean & 20 & 0.0001 & 0.0 & 0.25 & 0.1 & 0.0 & 0.6 & 8 \\
    \bottomrule
  \end{tabular}
  }
\end{table}

\begin{table}
\caption{Additional experiment results on homophilic and large-scale graphs.}
  \label{tab:additional_exp}
  \setlength{\tabcolsep}{9pt}
  \resizebox{\linewidth}{!}{
  \begin{tabular}{cccc}
\toprule
\textbf{Methods} & Facebook & Wiki & ogbg-products \\
\midrule
BGRL & 89.71±0.35 & 79.02±0.13 & 78.59±0.02 \\
CCA-SSG & 89.45±0.60 & 78.85±0.32 & 75.27±0.05 \\
GraphMAE & 89.54±0.36 & 78.94±0.48 & 78.89±0.01 \\
GraphMAE2 & 88.49±0.43 & 78.84±0.44 & \textbf{81.59±0.02} \\
\midrule
GraphPAE & \textbf{91.46±0.23} & \textbf{79.32±0.29} & 79.10±0.02 \\
\bottomrule
\end{tabular}
}
\end{table}





\end{document}